# Simple 3D Pose Features Support Human and Machine Social Scene Understanding

Wenshuo Qin & Leyla Isik; Johns Hopkins University

## Abstract

Humans effortlessly recognize social interactions from visual input, yet the underlying computations remain unknown, and social interaction recognition challenges even the most advanced deep neural networks (DNNs). Here, we hypothesized that humans rely on 3D visuospatial pose information to make social judgments, and that this information is largely absent from most vision DNNs. To test these hypotheses, we used a novel pose and depth estimation pipeline to automatically extract 3D body joint positions from short video clips. We compared the ability of these body joints to predict human social judgments in the videos with embeddings from over 350 vision DNNs. We found that body joints predicted social judgments better than most DNNs. We then reduced the 3D body joints to an even more compact feature set describing only the 3D position and direction of people in the videos. We found that this minimal 3D feature set, but not its 2D counterpart, was necessary and sufficient to explain the prediction performance of the full set of body joints. These minimal 3D features also predicted the extent to which DNNs aligned with human social judgments and significantly improved their performance on these tasks. Together, these findings demonstrate that human social perception depends on simple, explicit 3D pose information.

## Introduction

Humans can instantly and effortlessly recognize whether two people are socially interacting based on visual input. This capacity is a core human ability, and growing research in psychology and neuroscience suggests that it depends on extracting visuospatial features, such as relative positions, directions, and motion of bodies, which provide the building blocks for higher-level inferences about others' beliefs and goals (McMahon & Isik, 2023; Papeo, 2020; Scholl & Gao, 2013). Many of these abilities are present in infancy (Geraci et al., 2025; Powell & Spelke, 2018) and are also shared with nonhuman primates (Krupenye & Hare, 2018).

At the same time, modern deep neural networks (DNNs) have made rapid advances in recognizing objects, scenes, and actions. DNNs trained on large-scale image and video datasets can match human accuracy on various vision tasks, including object categorization and scene captioning (Caron et al., 2021; Chen et al., 2020a; He et al., 2015; Radford et al., 2021). Yet these massively trained vision DNNs still cannot model many aspects of human social vision understanding (Lake et al., 2017; Netanyahu et al., 2021; Shu et al., 2021). For example, a recent study showed that while pretrained DNNs align with human judgments of physical and scene features, they diverge markedly from human judgments of social features, such as identifying whether two people are facing one another or engaged in a social interaction (Garcia et al., 2025). This suggests that the representations learned by current vision DNNs, including state-of-the-art (SOTA) image and video models trained on various supervised and self-supervised tasks, miss key information that humans rely on to recognize social interactions.

A likely reason that DNN models fail is that humans rely on the visuospatial features from human bodies, such as gaze direction and physical contact, to make social judgments (McMahon & Isik, 2023), that is not captured by most pretrained vision models. Recent work has shown that cognitively based models of these abilities, recurrent and graph neural networks that get explicit information about agents and their visuospatial gaze information, can rival these SOTA models with a fraction of the training data and learnable parameters (Malik & Isik, 2023; Qin et al., 2025), supporting the idea that the spatial configuration of bodies is critical for human social judgments. However, these prior models still have three major limitations: (1) they rely on manually annotated inputs, (2) they focused on single gaze feature that only captures gaze-related social interaction while ignoring other social dimensions, such as physical interaction (Qin et al., 2025) and (3), like other neural network approaches, they are difficult to interpret and understand (Bowers et al., 2023; McCloskey, 1991).

In the current study, we take the same cognitive inspiration that the visuospatial configurations of bodies are critical for human social interaction judgments, while addressing the key limitations of prior work. We develop an image-computable framework to extract 3D full-body information directly from visual input, making our approach image-computable, holistic, and interpretable. We then use our modeling framework to test two central hypotheses: (1) that humans rely on 3D pose information to make social interaction judgments, and (2) that this information is missing in most modern vision DNNs.

To test these hypotheses, we use a dataset of short naturalistic video clips depicting two people engaged in everyday actions. Each video clip is annotated with human ratings along dimensions that span from low-level scene perception to higher-level social interaction ratings (McMahon et al., 2023). To extract pose information from these videos, we leverage recent advances in human pose estimation for accurate, multi-person spatial poses from single-view videos. We combine state-of-the-art pose (Goel et al., 2023) and depth estimation (Sun et al., 2022) models to extract 3D body joints from the videos and evaluate how well these body joints predict human social ratings (Figure 1). We compare the performance of these pose features with embeddings from over 350 vision DNNs, which include both image- and video-based models that span a range of architectures and training objectives. Next, motivated by cognitive theories proposing that social perception relies on abstracted, task-relevant visuospatial information (McMahon & Isik, 2023; Zhou et al., 2019), we ask whether the social information captured by the dense 3D body joints can be explained by a set of more interpretable features describing only each agent’s position and facing direction. Finally, we ask to what extent off-the-shelf vision DNNs encode these visuospatial pose features, and whether their performance can be improved by integrating pose features with their embeddings.

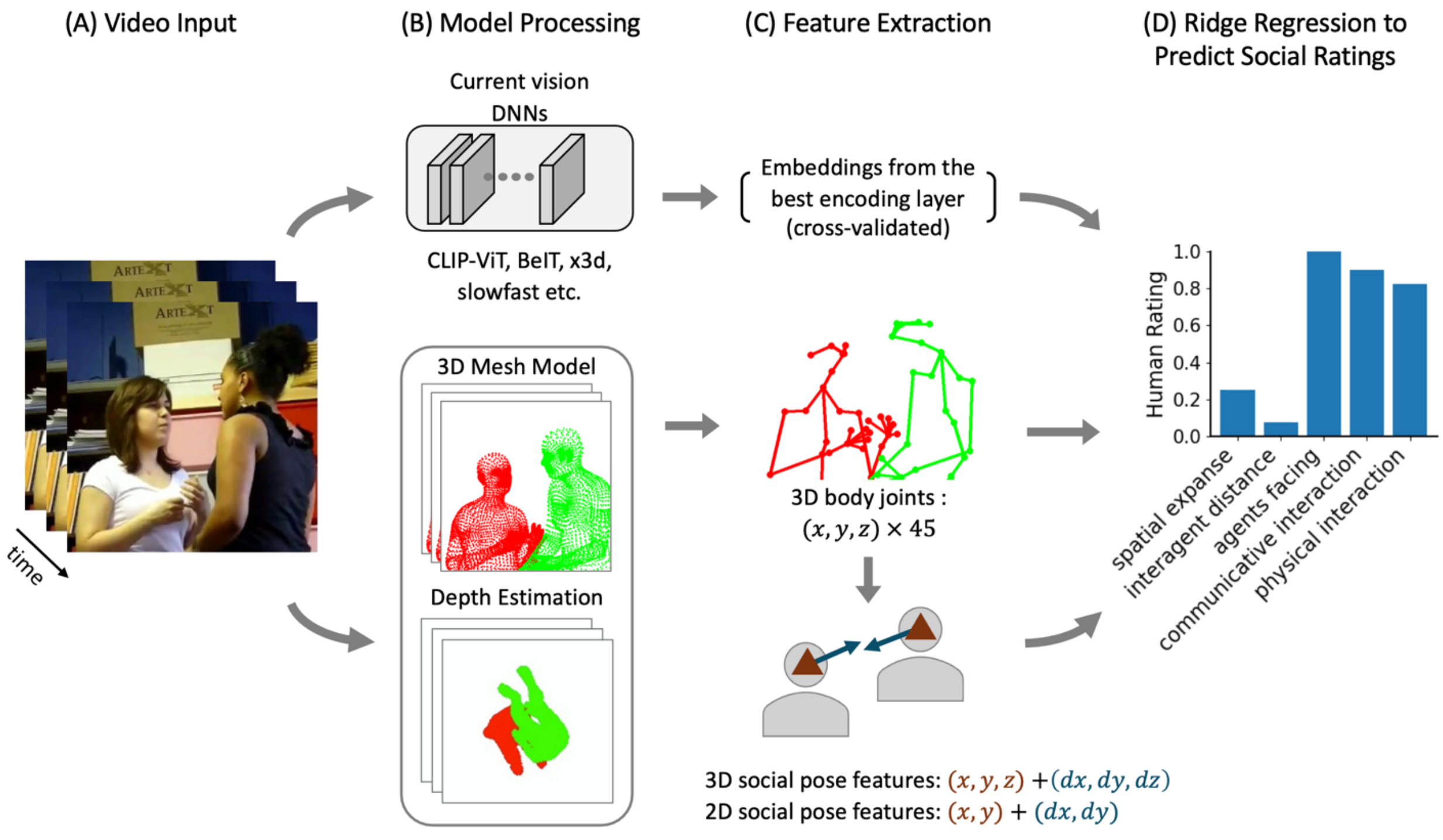

**Figure 1: Overview of the pose-based human rating prediction pipeline.** (A) Video frames from a naturalistic dataset are processed through (B) either pretrained vision models (e.g., CLIP-ViT, BeiT, x3d, slowfast) to extract learned visual embeddings (top), or the proposed 3D pose pipeline that combines a 3D mesh model and a pose depth estimation model for reconstructing the body mesh with accurate depth estimation (bottom). (C) Feature extraction yields embeddings from the best encoding layer of each vision model (determined via nested cross-validation, top) or 45 3D body joints (x, y, z) for each person (middle). From these joints, we derive compact 3D social pose features capturing each agent's position (x, y, z) and facing direction (dx, dy, dz) in 3D space and their 2D projections (x, y) + (dx, dy) (bottom). (D) Separate ridge regressions map each feature set to five human-rated behavioral dimensions: spatial expanse (scene size), interagent distance (distance between people in the video), agents facing (the extent to which people are oriented towards or away from each other), communicative interaction (whether people are communicating), and physical interaction (whether people are acting together physically). Each feature set was evaluated on the Pearson correlation score between the model-predicted ratings and the actual human ratings.

# Results

## Explicit 3D body joints outperform most vision DNN embeddings in predicting human judgments

We compared the ability of layer-wise embeddings from different off-the-shelf vision DNNs to predict human ratings of short video clips (Figure 1). We used a dataset of 250 three-second, silent videos depicting two people engaged in natural, everyday actions, drawn from the *Moments in Time* action recognition dataset (McMahon et al., 2023; Monfort et al., 2020), each annotated with behavioral ratings on spatial expanse (scene size), interagent distance (distance between two people), agents facing (whether or not people are facing each other), communicative interaction (whether or not people are communicating), and physical interaction (whether or not people are acting together). Off-the-shelf vision models included static image networks (e.g., BeIT, SimCLR) (Bao et al., 2021; Chen et al., 2020b), language-aligned multimodal models (e.g., CLIP-ViT) (Radford et al., 2021), and video models (e.g., X3D, TimeSformer) (Bertasius et al., 2021; Feichtenhofer, 2020) (for a full list, see Supplemental Table 1). We first replicated prior findings that, while most pretrained vision models can match human ratings of scene features (i.e., spatial expanse), these models still underperform in matching human social ratings, such as interagent distance, agents facing, communicative interaction, and physical interaction. Prior work has shown that this shortcoming was true, regardless of architecture, training data, or training objective (Garcia et al., 2025).

We next asked whether 3D pose information could predict human social ratings in the videos. We developed a pipeline to extract full-body joints for the two people, defined using the 3D 45-joint SMPL-X (Skinned Multi-Person Linear eXpressive) representation of body, face, hand, and foot landmarks, averaged across all 90 frames for each video for simplicity. We then tested the predictive performance of 3D body joints and found that they consistently outperformed the mean prediction accuracy of vision models across all five behavioral ratings (Fig. 2). For spatial expanse, 3D body joints outperformed the average vision model by a correlation of 0.06, exceeding 93% of all models tested. For interagent distance, 3D body joints showed an advantage of 0.05, exceeding 74% of models. For agents facing, 3D body joints outperformed the average model by 0.25, exceeding 99% of models. For communicative interaction, 3D body joints outperformed by 0.07, exceeding 76% of models. For physical interaction, 3D body joints outperformed by 0.27, exceeding 98% of models. The advantage of 3D body joints over vision DNN embeddings was statistically significant for predicting agents facing ($p = 0.0418$, two-tailed permutation test).

To understand whether the advantage of 3D body joints could also be seen in the internal representations of the model that generates these joints, we identified the best-performing embedding layer from 4D Humans (Goel et al., 2023) using the same pipeline as our other vision DNNs.  Surprisingly, this layer performed worse than the mean of standard vision DNNs and the explicit 3D body joints across all behavioral dimensions, highlighting the importance of the 3D pose outputs.

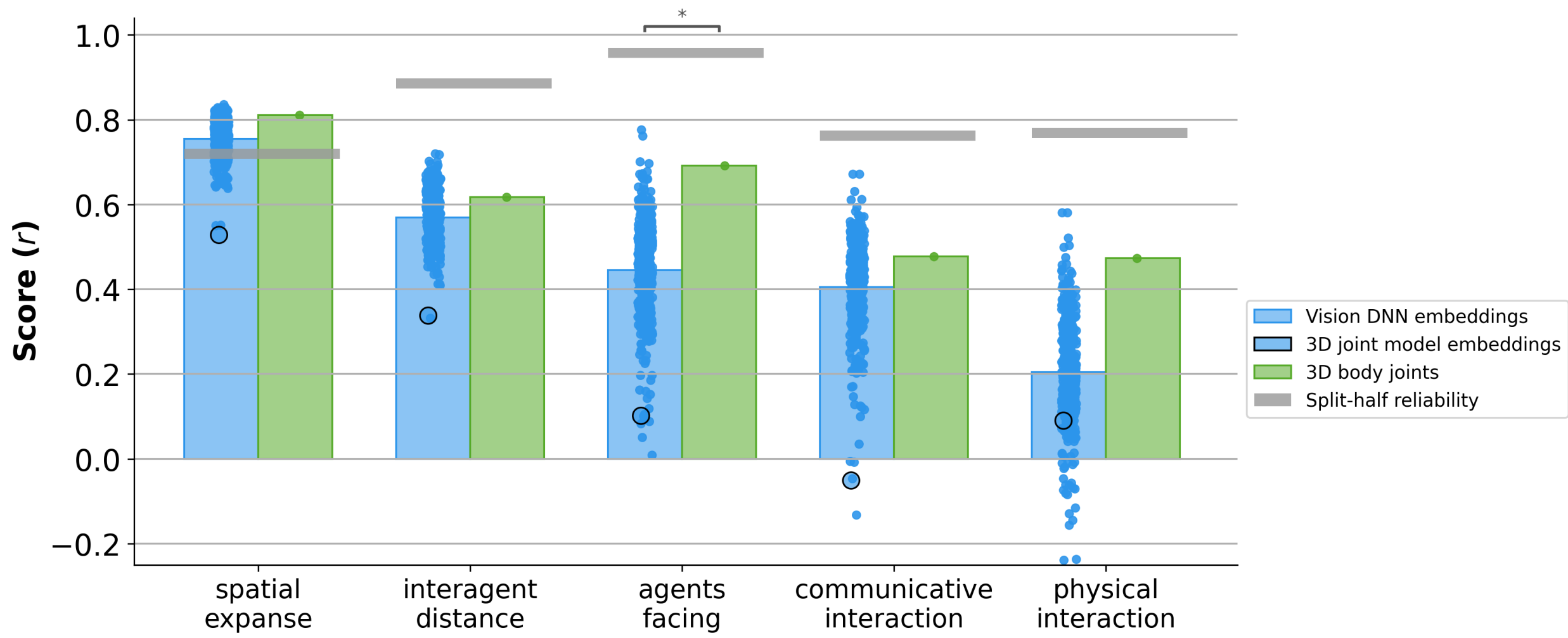

**Figure 2: Vision DNNs and 3D body joints predictions of human ratings.** Comparison of Vision DNNs (blue), 3D body joints (green), and the 3D joint model embeddings (circled blue) performance in predicting different human social scene ratings. Score is measured as the Pearson correlation ($r$) between the predicted and the true human ratings across five dimensions: spatial expanse, interagent distance, agents facing, communicative interaction, and physical interaction. The average performance was shown in bars. Each dot in the vision DNN embeddings shows the performance of the model's best layer (cross-validated) for each judgment. The larger circled blue dot represents the best layer performance of the 3D joint model. Gray horizontal bars indicate the split-half reliability of human ratings. Asterisks mark significance levels from two-tailed non-paired permutation tests (*$p < 0.05$).

## Simple 3D social pose features match dense 3D body joints in predicting human ratings

To identify which aspects of the 3D body joints drive social judgments, we designed a compact set of 3D social pose features capturing position and direction for each person, inspired by prior work emphasizing these dimensions as foundations for social interaction recognition (McMahon & Isik, 2023). These features consisted of position $(x, y, z)$ and direction $(dx, dy, dz)$, which are only 12-dimensional (6 dimensions × 2 agents) and simpler than the 270-dimensional full body joints (45 joints × 3 coordinates × 2 agents). To understand the contribution of 3D visuospatial information, we also tested the 2D counterparts of these features $(x, y, dx, dy)$.

We found the compact 3D social pose features performed nearly identically to the full set of 3D joints across all five ratings, with an average difference in correlation of only 0.03 across all five ratings (Fig. 3). In contrast, 2D social pose features showed a consistent reduction in performance relative to the full 3D joints, with an average difference of 0.29 in correlation across all five ratings. Consistent with these results, semi-partial correlation analyses (Supplementary Fig. S1) confirmed that the full 3D body joints contribute little additional prediction to human ratings above the simple 3D pose features.

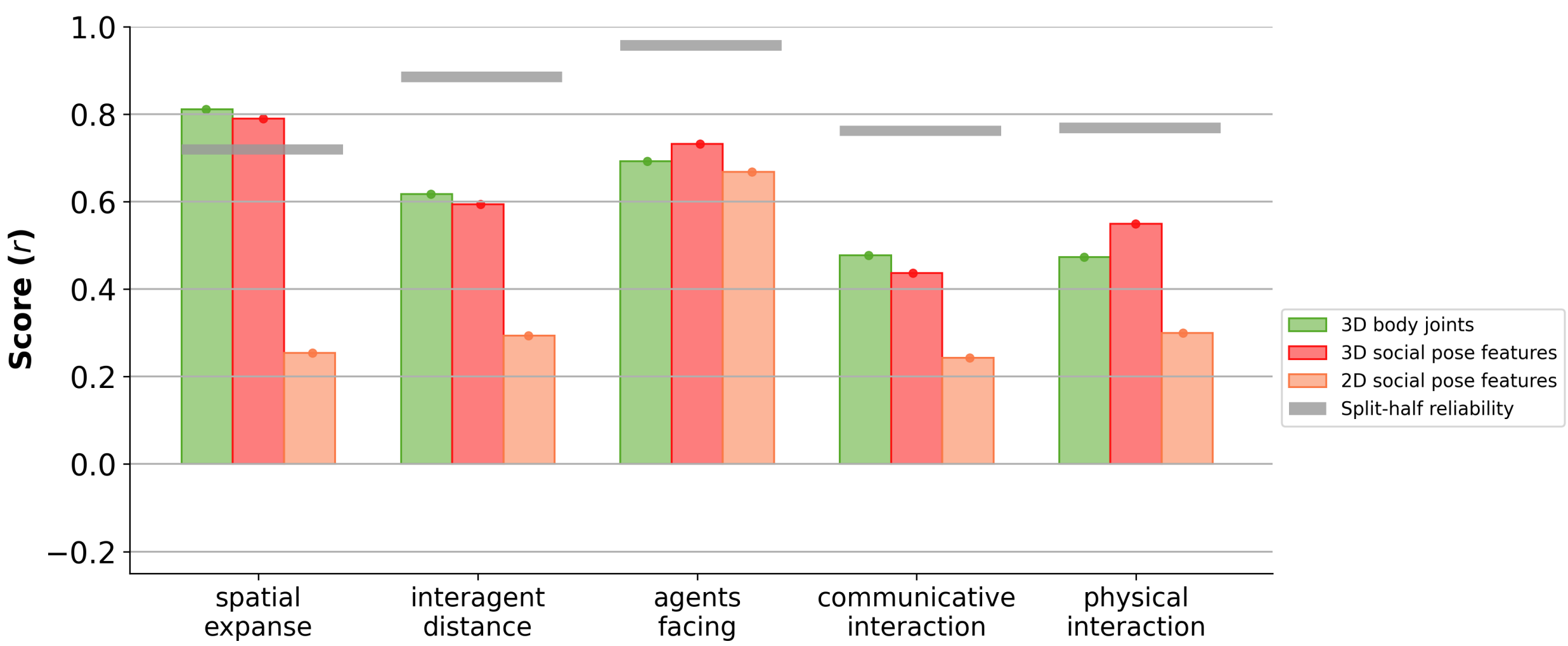

**Figure 3: Compact 3D social pose features predict human judgments as well as full 3D body joints.** Comparison of 3D body joints (green), 3D social pose features (red), and 2D social pose features (orange). Score is measured as the Pearson correlation ($r$) between the predicted and the true human ratings across five dimensions: spatial expanse, interagent distance, agents facing, communicative interaction, and physical interaction. Gray horizontal bars indicate the split-half reliability of human ratings.

## Off-the-shelf vision DNNs that encode 3D social pose features better predict human social ratings

Given the success of the 3D social pose features, we asked whether off-the-shelf vision DNNs that represent more 3D pose information in their learned embeddings can better predict human ratings of social interactions. To answer this question, we used the best model layer for each DNN (as chosen in training, see Methods) and evaluated how well that layer's embeddings predicted the 3D social pose features.

We found that models that better predicted 3D social pose features tended to achieve higher alignment with human social ratings but not scene ratings (Fig. 4). Specifically, we found significantly positive correlations for all social ratings: interagent distance ($r = 0.39$, $p = 0.0002$, two-tailed permutation test), agents facing ($r = 0.66$, $p = 0.0002$), communicative interaction ($r = 0.52$, $p = 0.0002$), and physical interaction ($r = 0.32$, $p = 0.0002$). Yet we found no clear trend for spatial expanse ($r = 0.00$, $p = 0.953$), a scene-centric rating.

We also found that models that better predicted 2D social pose features were more aligned with human social judgments: interagent distance ($r = 0.21$, $p = 0.0004$), agents facing ($r = 0.43$, $p = 0.0002$), communicative interaction ($r = 0.36$, $p = 0.0002$), and physical interaction ($r = 0.20$, $p =$ 0.0006). Yet there was a negative correlation with spatial expanse ($r$ = -0.20, $p$ = 0.0002). In addition, all correlations with 2D social pose feature prediction were significantly less than those with their 3D counterparts (permutation tests: $p$ = 0.0002 for spatial expanse, interagent distance, agents facing, and communicative interaction; and $p$ = 0.001 for physical interaction).

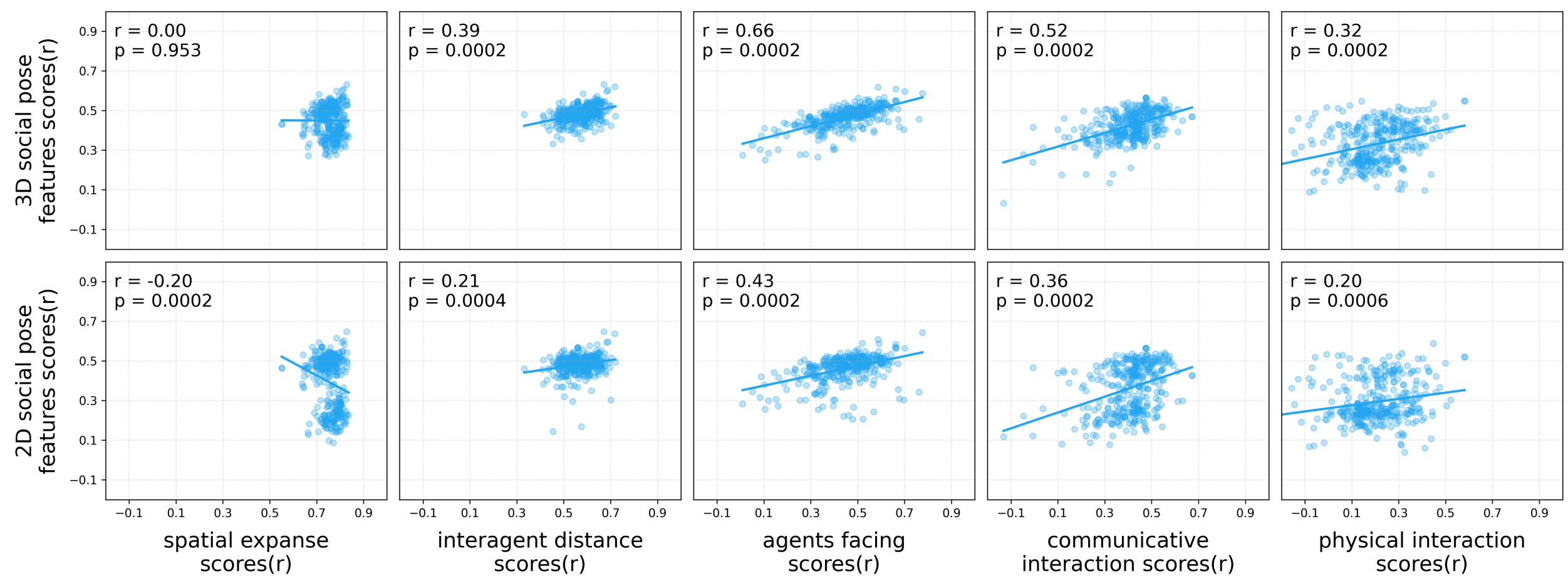

**Figure 4: Vision DNNs that better encode 3D social pose features better predict human social ratings.** Each scatterplot shows the relationship between vision DNN performance in predicting the human rating (x-axis) and predicting 3D or 2D social pose features (y-axis). Each dot represents one model's best layer, selected via cross-validation based on human rating prediction performance; this same layer is then evaluated for social pose feature encoding. Scores on the axes are measured as the Pearson correlations ($r$) between the predicted and the true human rating or social pose features. The regression line indicates the overall trend between the two prediction correlations, and the top left correlation coefficient ($r$) indicates the strength of this relationship.

## 3D social pose features augment vision DNN predictions of human ratings

Finally, we asked whether 3D social pose features provide overlapping or complementary information to modern vision models. To test this, we combined the 3D social pose features with the embeddings from a large set of vision DNNs using grouped ridge regression and evaluated their joint predictive power for human ratings.

Across all five behavioral ratings, adding 3D social pose features to vision DNN embeddings significantly improved prediction performance (Fig. 5; $p < 0.001$ for all ratings, paired one-tailed permutation test). The improvement was substantial and consistent: on average, correlations increased by 0.06 for spatial expanse (with 92% of models showing improvement), 0.08 for interagent distance (87% of models improved), 0.29 for agents facing (99% of models improved), 0.15 for communicative interaction (92% of models improved), and 0.08 for physical interaction (66% of models improved). These results show that simple 3D pose features provide information that is not captured in the learned embeddings of most modern pretrained models.

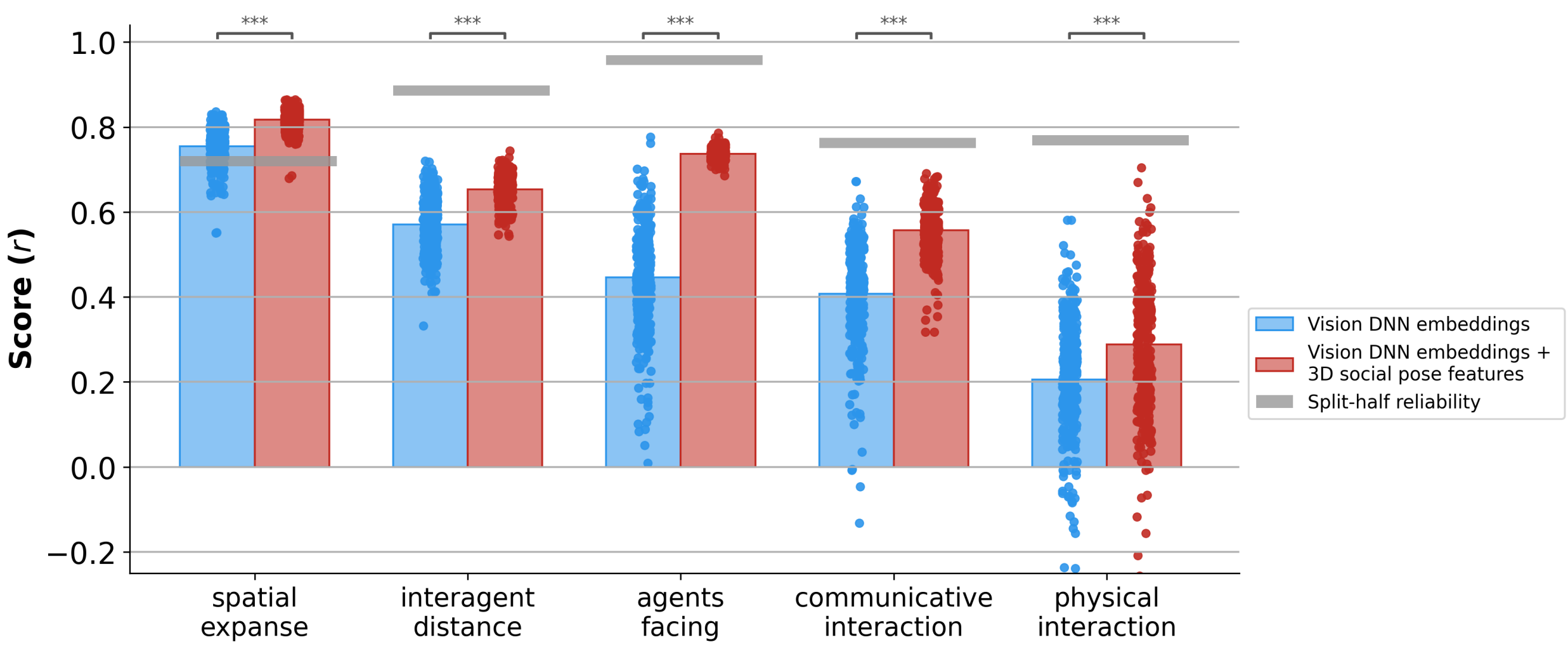

**Figure 5: 3D social pose features enhance vision DNN predictions of human ratings.** Comparison of Pearson correlations ($r$) between (1) Vision DNN embeddings alone and (2) Vision DNN embeddings combined with 3D social pose features. Each dot represents one model's best layer's prediction. Bars indicate group means. Gray horizontal bars indicate the split-half reliability of human ratings. Asterisks denote significance levels from one-tailed paired permutation tests (***$p < 0.001$).

## Discussion

In this study, we tested whether explicit 3D pose information can account for human social judgments and improve the representations learned in modern DNNs. Across multiple human social scene judgments, we found that simple 3D pose-based representations, both the full body 3D body joints and a compact set of 3D social pose features, rival or surpass the performance of modern vision models, and that this compact set of 3D pose features can improve off-the-shelf models' human alignment on social tasks. These findings suggest that much of human social scene understanding can be supported by explicit, low-dimensional 3D pose features, which are lacking in most modern vision DNNs.

Our results showed that 3D body joints consistently outperformed off-the-shelf vision models across all behavioral dimensions, with advantages for socially relevant features like agents facing and physical interaction. This suggests that critical social information captured by the explicit 3D body joint positions is not adequately captured in the learned embeddings of most vision models. Even the 4D Humans model (Goel et al., 2023), which is the basis of our 3D pose estimation, does not capture social information, suggesting it lacks explicit 3D body representations in its embedding space. These findings highlight a fundamental gap in the representations learned by current architectures with end-to-end training, regardless of the training tasks and architectures (Garcia et al., 2025).

Our study also demonstrates that the social information in the body joint configurations can be largely captured by only two simple visuospatial features: the 3D position and the facing direction of each person in the scene. This result aligns with cognitive theories suggesting that simple visuospatial features like distances and directions serve as foundational building blocks for higher-level social interaction perception (Hafri et al., 2013; Hafri & Firestone, 2021; McMahon & Isik, 2023; Papeo, 2020; Zhou et al., 2019). In contrast, 2D pose features lacking depth information consistently underperformed, emphasizing that humans rely on explicit 3D layout when recognizing social scenes.

Moreover, we found that models whose embeddings more strongly encoded 3D social pose features achieved higher alignment with human social judgments. Adding 3D social pose features to vision model embeddings also significantly improved prediction performance across all behavioral dimensions, suggesting that the lack of explicit 3D pose information is a widespread limitation in learned embeddings of modern vision models. These findings suggest that more human-like machine social understanding may require not only scaling up current architectures or training datasets but also incorporating explicit pose representations. This pose information may enable more sample-efficient learning and more interpretable social interaction recognition.

This study highlights both key visuospatial features for human social judgments and limitations of modern DNNs and opens the door to many future directions. First, our dataset was relatively small. Future work should incorporate longer, more diverse videos with wider distributions of spatial layouts and edge cases where pose configurations may be more ambiguous or vary in their correlations with higher-level judgments. Second, our pipeline averages body joint positions across frames, likely discarding rich dynamics, like motion synchrony and communicative exchanges. More sophisticated temporal modeling may be needed for longer videos and interactions distinguished by temporal dynamics. Third, while position and facing direction explain most linear pose-based variance in human ratings, higher-level social reasoning like motivations and intentions likely require processing beyond these shallow features. Finally, this study also raises questions about how the human brain represents these 3D pose features during social perception. Neuroimaging studies suggest regions of the extrastriate body area may compute spatial relationships between agents, and encode view-invariant, three-dimensional configurations of bodies (Abassi & Papeo, 2020; Downing et al., 2001; Gandolfo et al., 2024; Zhu et al., 2024), and transform these into higher-level social judgments in the superior temporal sulcus (McMahon et al., 2023). Future work could provide a unified framework to explain how 3D body information is used in downstream social tasks.

## Conclusion

Our findings reveal that human social scene understanding relies on explicit representations of 3D visuospatial pose information and can be supported by remarkably simple pose features. Contemporary vision DNNs that represent this 3D pose information more explicitly are more human-aligned social interaction judgments, and most models benefit from incorporating such representations. Together, these results help explain why vision DNNs continue to struggle with social recognition despite strong performance on other visual tasks and offer a path toward vision models that see the social world in a more human-like manner.

# Methods

## Behavior ratings and video dataset

We used a publicly available dataset of 250 three-second, silent videos depicting two people engaged in natural, everyday actions, drawn from the *Moments in Time* action recognition dataset (McMahon et al., 2023; Monfort et al., 2020). Each video was annotated with a comprehensive set of human behavioral ratings collected through large-scale online experiments. Participants rated each feature on a scale of 1-5, and ratings were collected and averaged for at least 10 participants.

For the present study, we focused on five rating dimensions that span a spectrum from low-level scene properties to high-level social understanding based on their importance in behavior and brain responses (McMahon et al., 2023). The first feature, spatial expanse, is a scene feature that captures the perceived size or openness of the environment. The next two features are "social primitives", visuospatial features that are often indicative of social interaction (McMahon & Isik, 2023): interagent distance reflects how physically close or far apart people appear within the scene, and agents facing measures the extent to which individuals are oriented toward each other. Communicative interaction captures whether agents seem to be engaged in an exchange of information or attention, such as talking, gesturing, or making eye contact, whereas physical interaction reflects whether they are engaged in direct bodily contact or coordinated joint actions, such as dancing or fighting. These dimensions provide a graded behavioral framework for evaluating how well computational models capture the visuospatial and social information humans extract from dynamic natural scenes.

The dataset was originally divided into 200 training videos and 50 test videos to facilitate cross-validated model prediction, and we maintained that pre-determined split for model evaluation. Due to occasional pose estimation failures, for instance, when one or both agents were heavily occluded, partially outside the frame, or captured in unusual body configurations, the pose model was unable to reliably extract two-person skeletons across all 90 frames for a subset of clips. To ensure high-quality and complete 3D pose data, we excluded these clips, resulting in a final set of 177 training videos and 47 test videos. To maintain consistency across all analyses, this same reduced set of videos was used throughout every stage of the study.

## Off-the-shelf vision models

We adopted the same set of vision models used in other recent NeuroAI benchmarking research (Conwell et al., 2024; Garcia et al., 2025), evaluating how diverse modern vision models capture social and spatial information from naturalistic videos. This model set included 351 image-based and video-based architectures spanning a range of training datasets and tasks, and architectural designs.

The image models encompassed convolutional and transformer-based architectures, such as ResNets, EfficientNets, Swin Transformers, Vision Transformers (ViTs), and CLIP variants, sourced from open repositories including Torchvision (TorchVision maintainers and contributers, 2016), PyTorch Image Models (Timm) (Wightman, 2019), VISSL (Goyal et al., 2021), and OpenAI CLIP (Radford et al., 2021). These models were trained under a variety of objectives, including supervised classification, self-supervised contrastive learning, and multimodal vision–language alignment. In addition, we incorporated video models such as SlowFast (Feichtenhofer et al., 2019) and TimeSformer (Bertasius et al., 2021), which are specifically designed to capture temporal dynamics from short video sequences. See Supplemental Table 1 for a full list of models included.

Feature representations were extracted following the same procedures (Garcia et al., 2025) using DeepJuice (Conwell et al., 2024), a memory-efficient Python toolkit for large-scale model benchmarking. For image models, we sampled seven evenly spaced frames from each 3-second clip, processed each frame independently, and averaged their activations to obtain one embedding per video. Video models processed the full temporal sequence directly (downsampling frames based on each model's preprocessing pipeline) to yield a single video-level representation. To enable direct comparison across models with varying dimensionalities, all extracted features were reduced to a 4,732-dimensional space via GPU-optimized sparse random projection (SRP) ($\varepsilon = 0.1$), following the Johnson–Lindenstrauss lemma (Larsen & Nelson, 2014).

## Depth-aware 3D poses and social features extraction

### 3D body joints

We combined two state-of-the-art pose estimation algorithms to extract a set of full-body 3D body joints. We first used the 4D Humans model, a vision transformer-based HMR 2.0 (Human Mesh Recovery v2) model that regresses human-body parameters from frames and aggregates over time for stable, consistent estimations in videos (Goel et al., 2023). The output is the parameterization of SMPL-X (Skinned Multi-Person Linear eXpressive), a detailed parametric model that represents 3D body shape and articulated pose with expressive hands and faces (Pavlakos et al., 2019).

Several videos in our dataset contain babies and children. Because of the adult-biased shape prior and training data, 4D Humans systematically pushes child-sized bodies deeper into the scene; we correct only the global translation with the BEV ("Bird's-Eye View") model (Sun et al., 2022). BEV estimates depth in true 3D voxels and incorporates an age-aware SMPL+A prior, making it markedly better at recovering children's front-back position while staying robust to inter-person occlusion (Sun et al., 2022). Accordingly, for every detection, we replace the depth component of the 4D Humans camera translation with BEV's metric depth. This fusion keeps HMR 2.0's high-fidelity pose/shape while inheriting BEV's accurate, infant-aware depth ordering.

For every 90-frame clip, we used the depth-corrected SMPL-regressed 3D body joint coordinates, which output a 45-joint set including the canonical 24 SMPL body joints and 21 additional face, hand, and foot landmarks. We used the averaged 3D joint coordinates across 90 frames to predict human ratings.

### 3D social pose features

To identify which components of full-body joints drive predictions of human ratings, we designed a smaller, more interpretable set of 3D social pose features explicitly aimed at capturing the social information present in full joints. Inspired by McMahon and Isik (2023), which emphasizes distance and facing direction between people as foundations for social interaction recognition, we extract face position and direction for each person from full body joints using simple, linear operations described below (Fig. 1C).

To construct the representation, we defined the position as the midpoint between the two eye joints and computed a unit direction vector by averaging the head-center–to-nose and neck-to-nose vectors, providing both an origin and a direction consistent with the face plane and head–torso hinge. We computed both 2D and 3D versions of these position and direction features. The 3D version comprised $(x, y, z, dx, dy, dz)$, while the 2D version comprised $(x, y, dx, dy)$. As with the full body joints, for each 90-frame clip, these features were averaged across time for each of the two people. We then used these features to assess how much of the social information in the 3D body joints could be captured by this compact representation.

### Pose model embeddings

To examine how the 4D Humans model represents information relevant to social interaction understanding, we also analyzed its internal latent representations. Because 4D Humans is not natively supported by the DeepJuice feature extraction framework we used for other Vision DNNs, we implemented a custom procedure to manually extract intermediate activations from the model. Specifically, we targeted two key components: the HMR2 module, responsible for estimating 3D body geometry and camera parameters from individual frames, and the PoseTransformer module, which integrates temporal information across frames to produce smooth and consistent pose trajectories. For each video, we recorded the output of every sublayer. To make these activations comparable across layers and consistent with the Vision DNNs, we applied the same SRP dimensionality reduction (4732 dimensions) implemented with the DeepJuice framework.

## Encoding framework

### Ridge encoding model

To evaluate how well each model/feature set predicts human ratings, we implemented a ridge regression–based encoding framework with normalization and cross-validation (Garcia et al., 2025). We first select the best-performing layer as a representative for this model through a five-fold cross-validation repeated twice on the training set. Within each fold, both model features and behavioral ratings were z-score normalized using statistics computed from the training split and applied to both training and validation data. The regularization strength ($\alpha$) was selected from a logarithmically spaced grid ranging from $10^{-10}$ to $10^{10}$. After identifying the best-performing layer for each model, we recomputed normalization parameters on the full training data and evaluated the model on the held-out test set. Model performance was quantified as the Pearson correlation ($r$) between predicted and actual behavioral ratings on the test set.

For the 3D social pose features, we fit the ridge regression directly on the full training and test splits (there is no layer selection involved in the 3D features), applying the same normalization and $\alpha$-selection process.

### Grouped ridge

To jointly model the contributions of vision model embeddings and 3D social pose features, we used the GroupRidgeCV implementation from the Himalaya package (Dupré la Tour et al., 2022). This method extends standard ridge regression by applying separate L2 regularization penalties to predefined feature groups, allowing the model to learn optimal weighting for the two input spaces (the vision model embeddings and the 3D social pose features), and helps account for the differences in feature set dimensionality. The optimization was performed using a random-search solver, which sampled 200 candidate group-weight configurations ($\gamma$) from a Dirichlet distribution defined on the simplex (with concentration parameters of 0.1 and 1.0) and, for each configuration, evaluated a shared set of logarithmically spaced regularization strengths from $10^{-10}$ to $10^{10}$. All features and targets were z-score normalized before fitting. The grouped model was then trained on the full training set, and its predictive performance was evaluated as the Pearson correlation ($r$) between predicted and observed behavioral ratings on the held-out test set.

### Semi-partial correlation

To determine whether the concise and interpretable 3D social pose features captured all social information contained in the full 3D body joints, and whether each component of 3D social pose features (3D positions and 3D directions) was necessary, we performed a semi-partial correlation analysis. Specifically, we tested whether each feature set (3D positions, 3D directions, and their combination as 3D social pose features) can account for all the variance explained in the behavioral ratings that was predicted by the full 3D body joints, or whether the full joints contained additional unique information beyond what each feature provided.

To achieve this, we first used ridge regressions to predict the 3D body joints from each of the feature sets (3D positions alone, 3D directions alone, and the full 3D social pose features), and then subtracted the body joints predicted by these features from the original body joints. This step produced a residualized representation of the full body joints that retains only the information in the joints not linearly predictable by 3D position and/or direction. Then, we used ridge regression to predict the behavioral ratings from the residualized joint features, testing whether the joint information not captured by each feature set could still predict behavioral ratings. The ridge encoding model followed the same procedures as those described for the main analysis.

### Correlation between model encodings of 3D pose and human ratings

To study whether models that better encode 3D social pose features also better capture human social ratings, we correlated each model's encoding performance across the two types of predictions. We used the same DNN encoding scores on human ratings described above. We then measured how well the best-encoding layer selected for each human rating could predict the 3D social pose features using the same encoding pipelines. Therefore, for each best-performing layer, we obtained a pair of encoding scores, one for predicting human ratings and one for predicting the 3D social pose feature. We computed the correlations across all the models. The resulting relationship quantifies whether models' ability to represent 3D social pose features is related to their success in matching human ratings.

## Statistical analyses

Because the underlying distribution of our data is unknown, we used non-parametric statistics (permutation testing) for all statistical analyses.

We performed two-tailed, non-paired permutation tests to test whether vision DNN embeddings (multiple predictors) and 3D social pose features (a single predictor) predicted human ratings significantly differently from each other. The test statistic was defined as the difference between the mean correlation score of DNN embeddings and the correlation score of the 3D social pose features. Under the null hypothesis that both groups were equally predictive, we randomly shuffled the human ratings 5,000 times and recomputed the correlations of both classes to generate a null distribution of mean differences expected by chance. The two-tailed *p*-value was then calculated as the proportion of permuted differences whose absolute value exceeded that of the observed difference.

We used a one-tailed paired permutation test in the grouped ridge analysis to evaluate whether 3D pose features could augment the encoding scores of each DNN embedding. The test statistic was the mean paired difference in correlation scores (DNN alone vs DNN + 3D pose). Under the null hypothesis of no improvement, we randomly swapped 5,000 times the paired correlation scores between the two model versions and recomputed the mean difference. Then we calculated the *p*-value as the proportion of permuted differences greater than or equal to the observed one.

To test whether there was a significant relationship between how well DNNs predict social pose features (3D or 2D) and how well they predict human ratings, we conducted a two-tailed permutation test. The test statistic was defined as the Pearson correlation between social pose feature prediction scores and human rating prediction scores. Under the null hypothesis of no correlation between the two scores, we randomly shuffled the social pose feature prediction scores 5,000 times and recomputed the correlation to generate a null distribution of correlations expected by chance. The two-tailed p-value was calculated as the proportion of permuted correlations with absolute values greater than or equal to the absolute value of the observed correlation.

We used a one-tailed permutation test to assess whether the relationship between DNNs' social pose prediction accuracy and human rating prediction accuracy was stronger for 3D social pose features than for 2D social pose features. Let $r_{3D}$ denote the correlation between DNNs' 3D social pose prediction scores and their human rating prediction scores, and $r_{2D}$ denote the correlation between DNNs' 2D social pose prediction scores and their human rating prediction scores. The test statistic was the difference: $\Delta r = r_{3D} - r_{2D}$. Under the null hypothesis that 3D features provide no advantage over 2D features ($\Delta r \leq 0$), we performed 5,000 permutations. In each permutation, we randomly shuffled the 3D social pose prediction scores and the 2D social pose prediction scores across DNNs, breaking any relationship between pose prediction accuracy and human rating prediction accuracy. We then recomputed $r_{3D}$, $r_{2D}$, and $\Delta r$ from the shuffled data. The one-tailed p-value was calculated as the proportion of permuted differences greater than or equal to the observed difference.

## Acknowledgements

We thank Alan Yuille and Jiahao Wang for their assistance with the 3D mesh models, and Kathy Garcia for assisting with large-scale vision model benchmarking and code review. We also thank Mick Bonner, Manasi Malik, Yuanfang Zhao, Hannah Small, Kathy Garcia, and Abdul-Rahim Deeb for valuable discussions throughout this project. This research was supported by the National Institute of Mental Health (NIMH) under award R01MH132826 to Leyla Isik.

# Supplemental materials

## 3D social pose features capture most social information from full 3D body joints

To determine whether our concise 3D social pose features (the combination of 3D positions and 3D directions) captured all the socially relevant information contained within the full body 3D body joints, we performed a semi-partial correlation analysis (see Methods). This analysis tested whether the full 3D joints contained any unique predictive information beyond what was already captured by our feature sets.

We first considered the two components of our social pose features, 3D positions and directions, separately. When partialling out only the 3D positions (Figure S1), the semi-partial correlations for spatial expanse ($r = 0.09$) and interagent distance ($r = 0.20$) show a clear decrease, but the correlation for agents facing ($r = 0.75$), communicative interaction ($r = 0.28$), and physical interaction ($r = 0.32$) are only minimally impacted. When partialling out only the 3D directions (Figure S1), the semi-partial correlation did not change across all dimensions ($r = 0.798$ for spatial expanse, $r = 0.634$ for interagent distance, $r = 0.682$ for agents facing, $r = 0.437$ for communicative interaction, and $r = 0.458$ for physical interaction). The remaining predictive power in both residualized conditions demonstrated that neither 3D positions nor 3D directions alone was sufficient to capture all socially relevant information from the joints.

However, when the complete 3D social pose features (combining both 3D positions and 3D directions) were partialled out from the 3D body joints, the unique predictive power of the remaining joint information showed dramatic drops across all five human ratings ($r = 0.070$ for spatial expanse, $r = 0.241$ for interagent distance, $r = 0.081$ for agents facing, $r = 0.035$ for communicative interaction, and $r = 0.051$ for physical interaction).

This result demonstrates that our concise 3D social pose features are both necessary (as individual components were insufficient) and largely sufficient (as the combined features left little remaining variance to be explained). The full, high-dimensional 3D joint data contains minimal unique, linearly decodable information relevant to these ratings beyond what is already captured by our combined 3D positions and 3D directions.

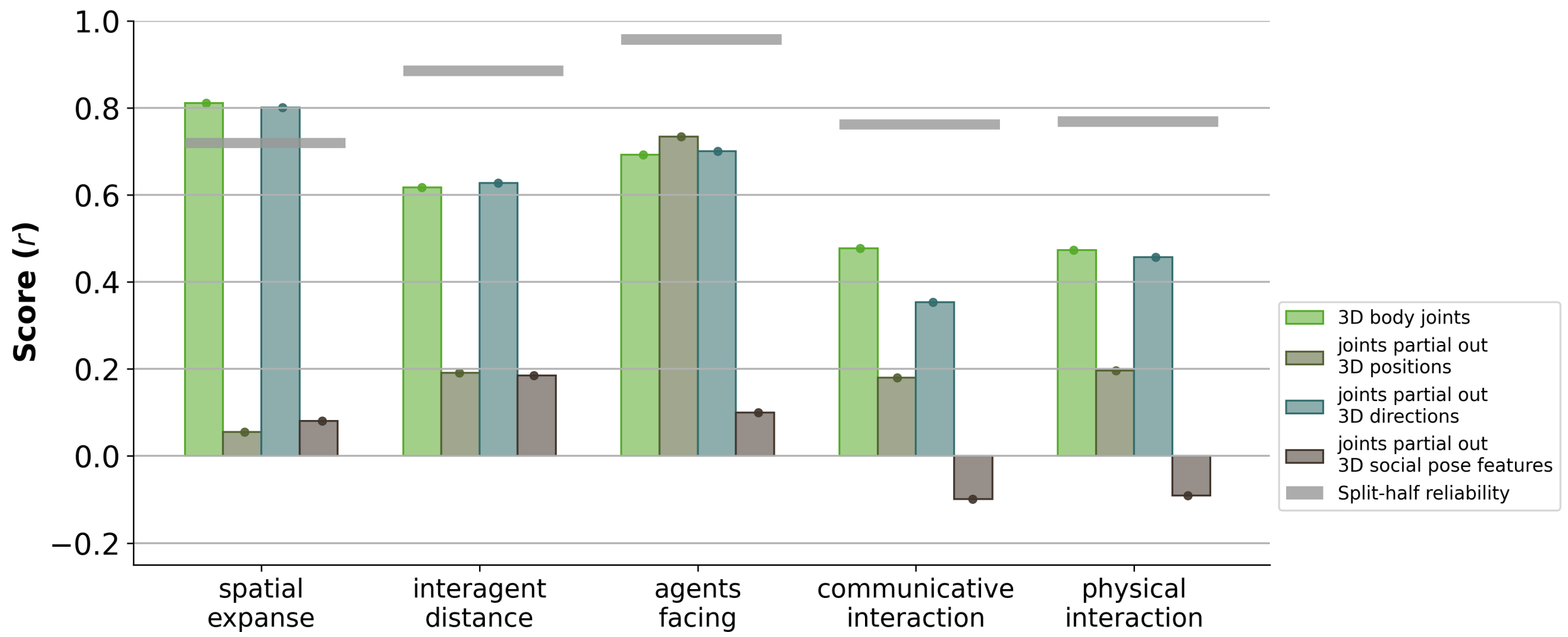

**Figure S1. 3D social pose features capture most social information from 3D body joints.** The bar chart displays the predictive performance (Pearson's *r*) for the full 3D body joints (light green) and the semi-partial correlation for residualized joint features, partialling out 3D positions (olive), 3D directions (teal), and the combination (dark brown). The dramatic drop in performance when the combination of position and direction is partialled out indicates that the full body 3D body joints contain little additional unique social information beyond the compact combination of 3D social pose features. Grey horizontal bars indicate the split-half reliability.

## List of vision DNNs tested

| model name | model type | spatial expanse | interagent distance | agents facing | communicative interaction | physical interaction |
|---|---|---|---|---|---|---|
| clip_vitl14 | image | 0.8089 | 0.7176 | 0.7763 | 0.6314 | 0.3346 |
| clip_rn50x4 | image | 0.8278 | 0.6717 | 0.6973 | 0.53 | 0.5216 |
| clip_vitb32 | image | 0.7556 | 0.6878 | 0.7621 | 0.5419 | 0.4581 |
| clip_rn101 | image | 0.7848 | 0.6831 | 0.5918 | 0.5188 | 0.447 |
| timm_convnext_large_in22ft1k | image | 0.7506 | 0.6505 | 0.6609 | 0.6724 | 0.2791 |
| timm_convnext_large | image | 0.7506 | 0.6505 | 0.6609 | 0.6724 | 0.2791 |
| timm_deit3_base_patch16_384_in21ft1k | image | 0.8289 | 0.6075 | 0.6056 | 0.576 | 0.3739 |
| timm_beitv2_base_patch16_224 | image | 0.7861 | 0.6293 | 0.701 | 0.4882 | 0.3759 |
| timm_convnext_small | image | 0.6986 | 0.5897 | 0.5889 | 0.5519 | 0.4594 |
| timm_deit3_base_patch16_224_in21ft1k | image | 0.8299 | 0.6927 | 0.6113 | 0.4775 | 0.2741 |
| timm_deit_small_distilled_patch16_224 | image | 0.7813 | 0.7205 | 0.5627 | 0.41 | 0.4066 |
| timm_deit3_large_patch16_384_in21ft1k | image | 0.7746 | 0.6521 | 0.6135 | 0.5325 | 0.3066 |
| torchvision_resnet50_imagenet1k_v2 | image | 0.7135 | 0.6175 | 0.6445 | 0.5489 | 0.3533 |
| timm_convnext_large_in22k | image | 0.7501 | 0.6328 | 0.6644 | 0.6114 | 0.2165 |
| timm_gernet_m | image | 0.7645 | 0.5981 | 0.6111 | 0.5571 | 0.3378 |
| timm_beit_large_patch16_384 | image | 0.8143 | 0.7021 | 0.6121 | 0.3498 | 0.3796 |
| timm_deit3_huge_patch14_224_in21ft1k | image | 0.7827 | 0.6467 | 0.5848 | 0.5473 | 0.2952 |
| timm_convnext_base_in22ft1k | image | 0.8104 | 0.6588 | 0.3108 | 0.4839 | 0.5809 |
| timm_convnext_base | image | 0.8104 | 0.6588 | 0.3108 | 0.4839 | 0.5809 |
| clip_vitb16 | image | 0.7587 | 0.6338 | 0.6726 | 0.4076 | 0.3687 |
| timm_deit3_small_patch16_224_in21ft1k | image | 0.7277 | 0.6177 | 0.6197 | 0.4813 | 0.3928 |
| timm_lambda_resnet26t | image | 0.7114 | 0.683 | 0.5983 | 0.4834 | 0.3542 |
| torchvision_regnet_x_1_6gf_imagenet1k_v1 | image | 0.7869 | 0.566 | 0.5204 | 0.4735 | 0.4755 |
| timm_dm_nfnet_f1 | image | 0.7363 | 0.6204 | 0.5701 | 0.574 | 0.2659 |

| timm_edgenext_x_small | image | 0.7984 | 0.6312 | 0.5029 | 0.4619 | 0.3692 |
|---|---|---|---|---|---|---|
| torchvision_mnasnet1_0_imagenet1k_v1 | image | 0.7934 | 0.6026 | 0.5335 | 0.5162 | 0.3025 |
| torchvision_regnet_x_800mf_imagenet1k_v1 | image | 0.7896 | 0.55 | 0.5769 | 0.5841 | 0.244 |
| timm_deit3_small_patch16_384_in21ft1k | image | 0.7346 | 0.6012 | 0.5988 | 0.4192 | 0.3879 |
| timm_convnext_base_384_in22ft1k | image | 0.7988 | 0.66 | 0.5962 | 0.2993 | 0.3829 |
| timm_convnext_base_in22k | image | 0.7566 | 0.5999 | 0.5026 | 0.4441 | 0.4168 |
| timm_beit_base_patch16_224 | image | 0.8288 | 0.6409 | 0.6313 | 0.4762 | 0.1413 |
| timm_deit3_large_patch16_384 | image | 0.7725 | 0.6467 | 0.5097 | 0.3796 | 0.4029 |
| timm_legacy_seresnet50 | image | 0.8022 | 0.6505 | 0.5472 | 0.429 | 0.2741 |
| slip_vit_b_yfcc15m | image | 0.7981 | 0.6605 | 0.4841 | 0.4626 | 0.2939 |
| x3d_xs | video | 0.7218 | 0.7015 | 0.5197 | 0.3259 | 0.4216 |
| timm_deit3_large_patch16_224 | image | 0.7904 | 0.5462 | 0.5178 | 0.5053 | 0.3288 |
| timm_cait_xxs36_224 | image | 0.8049 | 0.6501 | 0.4652 | 0.3268 | 0.4379 |
| timm_mixer_b16_224_miil | image | 0.7836 | 0.5438 | 0.6205 | 0.4051 | 0.328 |
| vicreg_resnet50_alpha0p75 | image | 0.7549 | 0.6162 | 0.3211 | 0.4895 | 0.499 |
| timm_edgenext_base | image | 0.7866 | 0.5993 | 0.5952 | 0.421 | 0.2736 |
| timm_legacy_seresnext50_32x4d | image | 0.7743 | 0.6547 | 0.4625 | 0.4463 | 0.3369 |
| timm_gluon_resnet101_v1c | image | 0.7346 | 0.6114 | 0.4201 | 0.5712 | 0.3359 |
| timm_convnext_tiny | image | 0.7955 | 0.6456 | 0.5452 | 0.4651 | 0.2208 |
| timm_cait_s24_224 | image | 0.8094 | 0.6792 | 0.4569 | 0.4174 | 0.308 |
| timm_convnext_pico | image | 0.7134 | 0.6322 | 0.4344 | 0.4474 | 0.443 |
| timm_jx_nest_tiny | image | 0.7954 | 0.6187 | 0.5137 | 0.4462 | 0.2963 |
| slip_vit_s_clip_yfcc15m | image | 0.7157 | 0.5926 | 0.3954 | 0.5498 | 0.4115 |
| torchvision_regnet_x_1_6gf_imagenet1k_v2 | image | 0.6936 | 0.633 | 0.5095 | 0.5539 | 0.275 |
| dino_resnet50 | image | 0.7369 | 0.5882 | 0.5626 | 0.5381 | 0.2334 |
| vissl_resnet50_pirl | image | 0.801 | 0.5931 | 0.596 | 0.5421 | 0.1242 |
| vissl_resnet50_swav | image | 0.801 | 0.5931 | 0.596 | 0.5421 | 0.1242 |
| vissl_resnet50_simclr | image | 0.801 | 0.5931 | 0.596 | 0.5421 | 0.1242 |
| timm_deit3_base_patch16_224 | image | 0.7711 | 0.5306 | 0.5734 | 0.4267 | 0.3539 |
| slowfast_r50 | video | 0.6455 | 0.5847 | 0.5685 | 0.5601 | 0.2969 |
| timm_convnext_small_in22k | image | 0.8114 | 0.5661 | 0.4246 | 0.4373 | 0.4144 |
| torchvision_swin_b_imagenet1k_v1 | image | 0.7516 | 0.5957 | 0.4419 | 0.4699 | 0.3912 |
| timm_gluon_resnext101_32x4d | image | 0.7298 | 0.6237 | 0.5137 | 0.5183 | 0.2619 |
| timm_gernet_s | image | 0.8253 | 0.6425 | 0.3869 | 0.4576 | 0.3211 |
| torchvision_convnext_small_imagenet1k_v1 | image | 0.7602 | 0.5442 | 0.4368 | 0.5032 | 0.3876 |
| torchvision_convnext_large_imagenet1k_v1 | image | 0.7316 | 0.6965 | 0.3301 | 0.4409 | 0.4271 |
| timm_mobilenetv3_large_100_miil | image | 0.7891 | 0.4879 | 0.6083 | 0.3781 | 0.3557 |
| timm_beitv2_large_patch16_224 | image | 0.7984 | 0.6597 | 0.6625 | 0.2026 | 0.2928 |
| x3d_s | video | 0.7835 | 0.6258 | 0.5224 | 0.4451 | 0.2387 |
| timm_convnext_nano | image | 0.805 | 0.6324 | 0.5833 | 0.4811 | 0.1107 |
| timm_eca_nfnet_l1 | image | 0.8009 | 0.5059 | 0.4996 | 0.2978 | 0.5038 |
| timesformer-base-finetuned-k400 | video | 0.739 | 0.5617 | 0.5821 | 0.4329 | 0.2909 |
| timm_convmixer_768_32 | image | 0.7756 | 0.6249 | 0.4804 | 0.3753 | 0.3492 |
| timm_gluon_resnet152_v1c | image | 0.7023 | 0.6442 | 0.4424 | 0.4935 | 0.3163 |
| timm_cs3darknet_l | image | 0.719 | 0.5842 | 0.4984 | 0.429 | 0.3679 |
| torchvision_efficientnet_v2_s_imagenet1k_v1 | image | 0.7638 | 0.5872 | 0.4948 | 0.47 | 0.2815 |
| timm_cs3darknet_m | image | 0.7357 | 0.622 | 0.6353 | 0.4462 | 0.1573 |
| dino_vitb8 | image | 0.811 | 0.6484 | 0.5523 | 0.3987 | 0.1805 |
| torchvision_regnet_y_3_2gf_imagenet1k_v2 | image | 0.7257 | 0.6313 | 0.4021 | 0.4428 | 0.389 |
| timm_cait_xs24_384 | image | 0.7608 | 0.6086 | 0.5349 | 0.464 | 0.2209 |
| dino_xcit_medium_24_p16 | image | 0.8137 | 0.6187 | 0.5338 | 0.4706 | 0.1516 |
| bit_expert_food | image | 0.7766 | 0.5994 | 0.5448 | 0.5146 | 0.151 |
| timm_gmlp_s16_224 | image | 0.7726 | 0.6225 | 0.4164 | 0.4129 | 0.362 |
| timm_deit3_huge_patch14_224 | image | 0.7706 | 0.6328 | 0.4031 | 0.5308 | 0.249 |
| timm_lamhalobotnet50ts_256 | image | 0.7282 | 0.6574 | 0.4391 | 0.463 | 0.2965 |
| torchvision_regnet_y_3_2gf_imagenet1k_v1 | image | 0.7731 | 0.5932 | 0.3716 | 0.415 | 0.429 |
| timm_mobilevitv2_150_384_in22ft1k | image | 0.7315 | 0.591 | 0.5826 | 0.4657 | 0.2109 |
| dino_vits16 | image | 0.7961 | 0.5373 | 0.6418 | 0.4192 | 0.184 |
| torchvision_efficientnet_b1_imagenet1k_v2 | image | 0.7672 | 0.5114 | 0.537 | 0.3549 | 0.3995 |
| timm_convit_base | image | 0.7896 | 0.6389 | 0.4944 | 0.3567 | 0.2901 |
| bit_expert_relation | image | 0.764 | 0.6015 | 0.5572 | 0.5148 | 0.1297 |
| vicreg_resnet50 | image | 0.7953 | 0.5523 | 0.5098 | 0.399 | 0.3062 |
| timm_convit_small | image | 0.7852 | 0.5857 | 0.6304 | 0.4463 | 0.1143 |

| torchvision_regnet_x_400mf_imagenet1k_v1 | image | 0.6876 | 0.5266 | 0.5064 | 0.5594 | 0.2792 |
|---|---|---|---|---|---|---|
| timm_dm_nfnet_f0 | image | 0.8287 | 0.5191 | 0.5027 | 0.5117 | 0.1956 |
| timm_halonet50ts | image | 0.7707 | 0.5389 | 0.4613 | 0.4639 | 0.3208 |
| torchvision_convnext_base_imagenet1k_v1 | image | 0.8279 | 0.6266 | 0.5003 | 0.2199 | 0.3762 |
| torchvision_swin_t_imagenet1k_v1 | image | 0.7804 | 0.6604 | 0.3223 | 0.4351 | 0.3526 |
| timm_deit3_medium_patch16_224 | image | 0.7488 | 0.5817 | 0.5366 | 0.4454 | 0.2373 |
| timm_ecaresnetlight | image | 0.8124 | 0.6348 | 0.5852 | 0.3955 | 0.1209 |
| torchvision_regnet_x_8gf_imagenet1k_v1 | image | 0.8024 | 0.5685 | 0.526 | 0.4123 | 0.2393 |
| timm_cs3sedarknet_x | image | 0.7975 | 0.5852 | 0.3159 | 0.4737 | 0.3744 |
| timm_cspresnext50 | image | 0.7068 | 0.5961 | 0.4828 | 0.5247 | 0.2316 |
| timm_cspresnet50 | image | 0.7698 | 0.5932 | 0.4088 | 0.4544 | 0.2995 |
| vicreg_resnet50_alpha0p9 | image | 0.7358 | 0.6036 | 0.5365 | 0.382 | 0.2663 |
| timm_deit3_large_patch16_224_in21ft1k | image | 0.8022 | 0.5821 | 0.5864 | 0.2696 | 0.2771 |
| timm_mobilenetv2_120d | image | 0.7505 | 0.6374 | 0.5906 | 0.3917 | 0.1429 |
| timm_dm_nfnet_f3 | image | 0.7881 | 0.4862 | 0.4283 | 0.4692 | 0.3379 |
| timm_dla60 | image | 0.8025 | 0.6242 | 0.417 | 0.4869 | 0.179 |
| torchvision_efficientnet_b1_imagenet1k_v1 | image | 0.7518 | 0.6215 | 0.4872 | 0.4361 | 0.21 |
| timm_legacy_seresnet101 | image | 0.8245 | 0.6235 | 0.4964 | 0.3211 | 0.2388 |
| timm_deit3_medium_patch16_224_in21ft1k | image | 0.7849 | 0.6183 | 0.4603 | 0.3502 | 0.2904 |
| timm_dpn92 | image | 0.7813 | 0.5733 | 0.3913 | 0.4452 | 0.3102 |
| timm_legacy_senet154 | image | 0.6804 | 0.5652 | 0.5131 | 0.4324 | 0.3077 |
| timm_levit_384 | image | 0.804 | 0.5437 | 0.5092 | 0.3984 | 0.2429 |
| timm_convnext_nano_ols | image | 0.7351 | 0.619 | 0.4787 | 0.4321 | 0.2315 |
| bit_expert_arthropod | image | 0.7717 | 0.615 | 0.434 | 0.4757 | 0.1959 |
| i3d_r50 | video | 0.6893 | 0.5585 | 0.3042 | 0.6122 | 0.3279 |
| bit_expert_flower | image | 0.7802 | 0.6004 | 0.4257 | 0.5264 | 0.1584 |
| timm_legacy_seresnet18 | image | 0.7546 | 0.5449 | 0.4428 | 0.4646 | 0.2823 |
| timm_gluon_inception_v3 | image | 0.7515 | 0.6034 | 0.4997 | 0.5016 | 0.1312 |
| slow_r50 | video | 0.7027 | 0.6435 | 0.4826 | 0.5279 | 0.1236 |
| timm_deit_tiny_distilled_patch16_224 | image | 0.7753 | 0.5786 | 0.4914 | 0.4268 | 0.2073 |
| timm_gluon_resnet50_v1b | image | 0.7259 | 0.5374 | 0.5104 | 0.4234 | 0.2784 |
| timm_edgenext_small | image | 0.797 | 0.5299 | 0.4362 | 0.5003 | 0.2084 |
| slip_vit_s_yfcc15m | image | 0.7352 | 0.6241 | 0.3496 | 0.5126 | 0.25 |
| timm_dla102x2 | image | 0.726 | 0.6423 | 0.3813 | 0.3583 | 0.3627 |
| timm_gluon_resnext50_32x4d | image | 0.6777 | 0.6435 | 0.453 | 0.4427 | 0.2538 |
| dino_xcit_small_12_p16 | image | 0.8256 | 0.5962 | 0.5035 | 0.391 | 0.1526 |
| timm_cait_xxs24_224 | image | 0.7872 | 0.5824 | 0.5531 | 0.3378 | 0.2062 |
| timm_levit_256 | image | 0.7598 | 0.6334 | 0.5589 | 0.3601 | 0.146 |
| timm_convnext_xlarge_in22k | image | 0.7815 | 0.6701 | 0.4171 | 0.4488 | 0.1378 |
| bit_expert_bird | image | 0.7765 | 0.6459 | 0.4431 | 0.4551 | 0.1335 |
| timm_cs3edgenet_x | image | 0.7224 | 0.6015 | 0.418 | 0.4485 | 0.2636 |
| clip_rn50 | image | 0.7098 | 0.6424 | 0.6778 | 0.431 | -0.0079 |
| bit_expert_instrument | image | 0.7695 | 0.6229 | 0.441 | 0.4761 | 0.1432 |
| timm_beit_large_patch16_224 | image | 0.8185 | 0.6405 | 0.525 | 0.4175 | 0.0502 |
| timm_botnet26t_256 | image | 0.6959 | 0.6764 | 0.5671 | 0.4458 | 0.0631 |
| timm_gluon_senet154 | image | 0.7297 | 0.5644 | 0.4607 | 0.436 | 0.2565 |
| dino_xcit_medium_24_p8 | image | 0.8166 | 0.654 | 0.5092 | 0.2706 | 0.1966 |
| timm_convnext_xlarge_in22ft1k | image | 0.7742 | 0.676 | 0.4312 | 0.4302 | 0.1333 |
| torchvision_swin_s_imagenet1k_v1 | image | 0.7827 | 0.5631 | 0.4251 | 0.3796 | 0.2943 |
| timm_dla60x | image | 0.7421 | 0.4712 | 0.4524 | 0.4172 | 0.3593 |
| timm_deit_small_patch16_224 | image | 0.7357 | 0.5909 | 0.4591 | 0.4277 | 0.2252 |
| timm_gluon_seresnext50_32x4d | image | 0.7044 | 0.6263 | 0.4606 | 0.334 | 0.312 |
| timm_halo2botnet50ts_256 | image | 0.7638 | 0.5997 | 0.5166 | 0.5417 | 0.0141 |
| timm_mobilevit_s | image | 0.7391 | 0.5147 | 0.6189 | 0.384 | 0.1762 |
| torchvision_regnet_x_800mf_imagenet1k_v2 | image | 0.7427 | 0.4815 | 0.5756 | 0.5719 | 0.0611 |
| timm_convmixer_1024_20_ks9_p14 | image | 0.7908 | 0.6078 | 0.4034 | 0.4062 | 0.2243 |
| timm_gluon_resnext101_64x4d | image | 0.7593 | 0.6274 | 0.4454 | 0.2864 | 0.3139 |
| torchvision_regnet_y_800mf_imagenet1k_v1 | image | 0.7351 | 0.537 | 0.3739 | 0.4618 | 0.3248 |
| torchvision_regnet_y_1_6gf_imagenet1k_v2 | image | 0.8032 | 0.4925 | 0.44 | 0.5224 | 0.1731 |
| torchvision_mobilenet_v3_large_imagenet1k_v1 | image | 0.8015 | 0.4971 | 0.3946 | 0.4185 | 0.3159 |
| torchvision_mnasnet1_3_imagenet1k_v1 | image | 0.6932 | 0.5376 | 0.5014 | 0.4913 | 0.2035 |
| timm_ecaresnet50d | image | 0.8182 | 0.4872 | 0.3948 | 0.4565 | 0.269 |
| timm_beit_base_patch16_384 | image | 0.7786 | 0.6621 | 0.5813 | 0.2299 | 0.1724 |

| timm_dla46x_c | image | 0.7421 | 0.5753 | 0.4705 | 0.4093 | 0.2252 |
|---|---|---|---|---|---|---|
| timm_deit_base_distilled_patch16_224 | image | 0.8032 | 0.4689 | 0.5316 | 0.3953 | 0.2227 |
| timm_mixer_b16_224_miil_in21k | image | 0.8015 | 0.554 | 0.3198 | 0.4304 | 0.3159 |
| torchvision_resnext101_32x8d_imagenet1k_v2 | image | 0.7092 | 0.6704 | 0.3667 | 0.5083 | 0.1634 |
| timm_convit_tiny | image | 0.7504 | 0.5993 | 0.5701 | 0.3566 | 0.1416 |
| bit_expert_object | image | 0.7715 | 0.6122 | 0.439 | 0.465 | 0.1277 |
| timm_convnext_tiny_in22ft1k | image | 0.7617 | 0.565 | 0.4912 | 0.4384 | 0.1574 |
| torchvision_mobilenet_v3_small_imagenet1k_v1 | image | 0.7403 | 0.5131 | 0.4941 | 0.3639 | 0.3017 |
| vissl_resnet50_deepclusterv2 | image | 0.7388 | 0.5837 | 0.4622 | 0.5423 | 0.086 |
| timm_deit3_small_patch16_224 | image | 0.7807 | 0.4951 | 0.5107 | 0.5225 | 0.1019 |
| timm_cs3darknet_focus_l | image | 0.7019 | 0.5189 | 0.5037 | 0.4309 | 0.2546 |
| torchvision_regnet_y_16gf_imagenet1k_v1 | image | 0.7523 | 0.4894 | 0.4866 | 0.4109 | 0.2683 |
| timm_mixer_b16_224 | image | 0.7497 | 0.5962 | 0.3957 | 0.3842 | 0.2809 |
| timm_mixer_b16_224_in21k | image | 0.7497 | 0.5962 | 0.3957 | 0.3842 | 0.2809 |
| timm_gluon_resnet50_v1d | image | 0.8017 | 0.5135 | 0.359 | 0.3691 | 0.3626 |
| torchvision_regnet_y_32gf_imagenet1k_v2 | image | 0.7402 | 0.5201 | 0.4323 | 0.3822 | 0.3308 |
| timm_dla102x | image | 0.7483 | 0.5859 | 0.475 | 0.4287 | 0.1656 |
| timm_cs3darknet_x | image | 0.7487 | 0.5354 | 0.5086 | 0.4163 | 0.1925 |
| torchvision_swin_v2_b_imagenet1k_v1 | image | 0.7535 | 0.5072 | 0.4343 | 0.3197 | 0.3837 |
| timm_levit_192 | image | 0.7742 | 0.4129 | 0.4963 | 0.546 | 0.1684 |
| torchvision_regnet_y_400mf_imagenet1k_v2 | image | 0.7348 | 0.4765 | 0.5207 | 0.3067 | 0.3581 |
| torchvision_shufflenet_v2_x1_5_imagenet1k_v1 | image | 0.7934 | 0.4378 | 0.5218 | 0.4207 | 0.2227 |
| timm_jx_nest_base | image | 0.7851 | 0.5698 | 0.4392 | 0.3652 | 0.237 |
| timm_mobilevitv2_150 | image | 0.7171 | 0.582 | 0.5419 | 0.3832 | 0.1668 |
| bit_expert_vehicle | image | 0.7666 | 0.6086 | 0.4512 | 0.4307 | 0.1335 |
| slip_vit_s_simclr_yfcc15m | image | 0.7746 | 0.5456 | 0.3739 | 0.4481 | 0.2473 |
| timm_cspdarknet53 | image | 0.6817 | 0.5432 | 0.445 | 0.5563 | 0.1621 |
| timm_densenet121 | image | 0.8072 | 0.5385 | 0.509 | 0.356 | 0.1766 |
| bit_expert_abstraction | image | 0.7647 | 0.614 | 0.4403 | 0.4436 | 0.121 |
| torchvision_resnet101_imagenet1k_v1 | image | 0.7571 | 0.6102 | 0.3132 | 0.5384 | 0.1634 |
| timm_gmixer_24_224 | image | 0.771 | 0.567 | 0.3812 | 0.4322 | 0.2306 |
| timm_halonet26t | image | 0.753 | 0.6144 | 0.5976 | 0.3358 | 0.0811 |
| torchvision_resnet34_imagenet1k_v1 | image | 0.7962 | 0.5937 | 0.4995 | 0.2635 | 0.2289 |
| torchvision_mobilenet_v2_imagenet1k_v2 | image | 0.7382 | 0.5901 | 0.3623 | 0.5305 | 0.1606 |
| timm_efficientformer_l1 | image | 0.7476 | 0.5033 | 0.3834 | 0.4063 | 0.3403 |
| timm_gluon_resnet50_v1c | image | 0.7392 | 0.5382 | 0.528 | 0.4904 | 0.0839 |
| timm_deit_base_distilled_patch16_384 | image | 0.7271 | 0.5959 | 0.4603 | 0.4143 | 0.181 |
| timm_hardcorenas_a | image | 0.7675 | 0.5539 | 0.5522 | 0.3712 | 0.1308 |
| torchvision_regnet_y_800mf_imagenet1k_v2 | image | 0.7578 | 0.5793 | 0.4173 | 0.4179 | 0.2016 |
| timm_hardcorenas_f | image | 0.7561 | 0.5292 | 0.477 | 0.3734 | 0.2358 |
| timm_cait_xxs36_384 | image | 0.7361 | 0.6453 | 0.5179 | 0.2735 | 0.1952 |
| torchvision_alexnet_imagenet1k_v1 | image | 0.7293 | 0.4987 | 0.3443 | 0.429 | 0.3623 |
| x3d_m | video | 0.715 | 0.5581 | 0.4617 | 0.4364 | 0.1922 |
| timm_mobilenetv3_small_075 | image | 0.7023 | 0.4729 | 0.4749 | 0.4413 | 0.2716 |
| timm_deit3_small_patch16_384 | image | 0.7528 | 0.6041 | 0.4397 | 0.46 | 0.1036 |
| timm_mnasnet_100 | image | 0.7555 | 0.4988 | 0.479 | 0.3969 | 0.2281 |
| torchvision_regnet_x_3_2gf_imagenet1k_v2 | image | 0.649 | 0.6495 | 0.4706 | 0.5223 | 0.064 |
| timm_dm_nfnet_f2 | image | 0.7816 | 0.6535 | 0.6721 | 0.3632 | -0.1157 |
| torchvision_mnasnet0_75_imagenet1k_v1 | image | 0.6924 | 0.5537 | 0.3551 | 0.4937 | 0.259 |
| timm_convnext_tiny_in22k | image | 0.7599 | 0.5338 | 0.3997 | 0.4497 | 0.2056 |
| torchvision_resnext50_32x4d_imagenet1k_v2 | image | 0.7338 | 0.6331 | 0.5033 | 0.43 | 0.048 |
| bit_expert_mammal | image | 0.7628 | 0.6143 | 0.4818 | 0.3028 | 0.1853 |
| timm_convnext_atto | image | 0.7573 | 0.5465 | 0.5479 | 0.425 | 0.0652 |
| vissl_resnet50_jigsaw_p100 | image | 0.7226 | 0.6083 | 0.374 | 0.4301 | 0.2047 |
| timm_gluon_resnet34_v1b | image | 0.7354 | 0.5529 | 0.4339 | 0.4282 | 0.1864 |
| torchvision_regnet_y_16gf_imagenet1k_v2 | image | 0.8025 | 0.5746 | 0.3672 | 0.4175 | 0.174 |
| vissl_resnet50_supervised | image | 0.7944 | 0.5608 | 0.3421 | 0.431 | 0.2062 |
| torchvision_shufflenet_v2_x1_0_imagenet1k_v1 | image | 0.7659 | 0.4924 | 0.3692 | 0.5145 | 0.1909 |
| timm_deit3_base_patch16_384 | image | 0.7928 | 0.5785 | 0.2467 | 0.4431 | 0.2715 |
| timm_dla169 | image | 0.688 | 0.5638 | 0.4199 | 0.3751 | 0.282 |
| timm_mobilenetv3_small_050 | image | 0.7077 | 0.4937 | 0.4526 | 0.511 | 0.1627 |
| timm_efficientformer_l7 | image | 0.7932 | 0.5933 | 0.3583 | 0.388 | 0.1926 |
| timm_mobilenetv3_large_100 | image | 0.7698 | 0.5315 | 0.422 | 0.5448 | 0.0573 |

| bit_expert_animal | image | 0.7687 | 0.6204 | 0.4648 | 0.3269 | 0.1436 |
|---|---|---|---|---|---|---|
| timm_deit_base_patch16_384 | image | 0.6968 | 0.532 | 0.4264 | 0.4596 | 0.2034 |
| timm_gluon_resnet101_v1s | image | 0.7556 | 0.6594 | 0.3772 | 0.5938 | -0.0705 |
| timm_convnext_tiny_hnf | image | 0.7371 | 0.5301 | 0.5311 | 0.4448 | 0.0722 |
| torchvision_ssd300_vgg16_coco_v1 | image | 0.8172 | 0.6585 | 0.4968 | 0.122 | 0.2197 |
| timm_mobilenetv2_050 | image | 0.7122 | 0.4812 | 0.3993 | 0.5127 | 0.2025 |
| dino_vits8 | image | 0.6414 | 0.5855 | 0.4134 | 0.4441 | 0.2226 |
| timm_levit_128 | image | 0.7982 | 0.4992 | 0.4664 | 0.4187 | 0.1214 |
| timm_deit_base_patch16_224 | image | 0.7244 | 0.5369 | 0.4038 | 0.4223 | 0.2161 |
| slip_vit_b_simclr_yfcc15m | image | 0.7786 | 0.5197 | 0.5082 | 0.2918 | 0.2026 |
| torchvision_convnext_tiny_imagenet1k_v1 | image | 0.793 | 0.6057 | 0.3446 | 0.4262 | 0.1302 |
| timm_convnext_tiny_384_in22ft1k | image | 0.7556 | 0.5602 | 0.4162 | 0.3542 | 0.2131 |
| torchvision_shufflenet_v2_x0_5_imagenet1k_v1 | image | 0.7532 | 0.51 | 0.2573 | 0.5074 | 0.2703 |
| torchvision_mobilenet_v3_large_imagenet1k_v2 | image | 0.7494 | 0.5056 | 0.4267 | 0.3853 | 0.2285 |
| timm_inception_v4 | image | 0.7506 | 0.479 | 0.5377 | 0.3845 | 0.1431 |
| torchvision_regnet_x_400mf_imagenet1k_v2 | image | 0.7256 | 0.5221 | 0.4122 | 0.336 | 0.2958 |
| timm_hardcorenas_c | image | 0.7854 | 0.4972 | 0.5458 | 0.3311 | 0.1299 |
| timm_edgenext_small_rw | image | 0.7994 | 0.5077 | 0.5089 | 0.3336 | 0.1391 |
| timm_mobilenetv2_110d | image | 0.7664 | 0.4854 | 0.34 | 0.4399 | 0.2471 |
| timm_haloregnetz_b | image | 0.7415 | 0.5096 | 0.4023 | 0.3559 | 0.2676 |
| timm_mobilevitv2_050 | image | 0.7624 | 0.555 | 0.4813 | 0.3386 | 0.1395 |
| torchvision_resnet18_imagenet1k_v1 | image | 0.7803 | 0.543 | 0.4728 | 0.3866 | 0.0905 |
| timm_gluon_resnet101_v1b | image | 0.7229 | 0.6223 | 0.5207 | 0.4298 | -0.0225 |
| vissl_resnet50_mocov2 | image | 0.7933 | 0.5943 | 0.3905 | 0.4162 | 0.0716 |
| timm_mobilenetv2_100 | image | 0.7642 | 0.5666 | 0.4456 | 0.3325 | 0.1563 |
| timm_gluon_resnet152_v1b | image | 0.7013 | 0.6136 | 0.3492 | 0.4133 | 0.1843 |
| timm_cait_s24_384 | image | 0.7051 | 0.6052 | 0.3252 | 0.3104 | 0.3116 |
| timm_bat_resnext26ts | image | 0.7417 | 0.5857 | 0.16 | 0.4581 | 0.3119 |
| timm_convmixer_1536_20 | image | 0.7604 | 0.4944 | 0.4794 | 0.397 | 0.1191 |
| torchvision_resnext101_64x4d_imagenet1k_v1 | image | 0.6604 | 0.5875 | 0.4834 | 0.2365 | 0.2819 |
| slip_vit_b_clip_yfcc15m | image | 0.7518 | 0.6038 | 0.3355 | 0.4 | 0.1579 |
| timm_cait_s36_384 | image | 0.7633 | 0.6635 | 0.4221 | 0.3462 | 0.0521 |
| timm_convnext_small_in22ft1k | image | 0.7637 | 0.5395 | 0.4682 | 0.3447 | 0.13 |
| timm_ecaresnet50t | image | 0.7832 | 0.4778 | 0.4326 | 0.4841 | 0.0678 |
| timm_hardcorenas_e | image | 0.7407 | 0.5263 | 0.4266 | 0.2954 | 0.256 |
| timm_efficientnet_b0 | image | 0.7885 | 0.5128 | 0.539 | 0.2505 | 0.1517 |
| torchvision_resnext50_32x4d_imagenet1k_v1 | image | 0.8236 | 0.6071 | 0.1015 | 0.4661 | 0.2406 |
| timm_mobilevit_xxs | image | 0.8066 | 0.5449 | 0.4466 | 0.3296 | 0.1093 |
| timm_mobilevit_xs | image | 0.7874 | 0.5761 | 0.4698 | 0.2662 | 0.1372 |
| timm_eca_resnet33ts | image | 0.7545 | 0.5112 | 0.493 | 0.2616 | 0.2144 |
| timm_mobilenetv3_large_100_miil_in21k | image | 0.771 | 0.5731 | 0.4246 | 0.4785 | -0.0131 |
| torchvision_resnet152_imagenet1k_v1 | image | 0.801 | 0.4099 | 0.4077 | 0.1697 | 0.4442 |
| torchvision_shufflenet_v2_x2_0_imagenet1k_v1 | image | 0.7131 | 0.5295 | 0.5435 | 0.5033 | -0.0571 |
| timm_gluon_resnet152_v1d | image | 0.7135 | 0.6344 | 0.2567 | 0.5372 | 0.0901 |
| timm_efficientnet_b2 | image | 0.7822 | 0.5356 | 0.4069 | 0.2725 | 0.2284 |
| timm_inception_resnet_v2 | image | 0.7791 | 0.6238 | 0.4585 | 0.2774 | 0.0865 |
| timm_ghostnet_100 | image | 0.7976 | 0.5858 | 0.5587 | 0.3405 | -0.0625 |
| torchvision_ssdlite320_mobilenet_v3_large_coco_v1 | image | 0.768 | 0.6451 | 0.272 | 0.4239 | 0.1107 |
| timm_mobilevitv2_150_in22ft1k | image | 0.731 | 0.477 | 0.3362 | 0.348 | 0.3272 |
| timm_hardcorenas_d | image | 0.7333 | 0.5149 | 0.4268 | 0.3583 | 0.1859 |
| timm_convnext_pico_ols | image | 0.7656 | 0.5545 | 0.4077 | 0.3986 | 0.0919 |
| timm_ig_resnext101_32x32d | image | 0.7226 | 0.5594 | 0.295 | 0.4756 | 0.1656 |
| timm_ig_resnext101_32x16d | image | 0.7226 | 0.5594 | 0.295 | 0.4756 | 0.1656 |
| timm_ig_resnext101_32x8d | image | 0.7226 | 0.5594 | 0.295 | 0.4756 | 0.1656 |
| timm_ig_resnext101_32x48d | image | 0.7226 | 0.5594 | 0.295 | 0.4756 | 0.1656 |
| timm_cait_xxs24_384 | image | 0.7873 | 0.5839 | 0.5085 | 0.4054 | -0.0681 |
| timm_levit_128s | image | 0.7554 | 0.5405 | 0.4035 | 0.4176 | 0.0976 |
| timm_eca_resnext26ts | image | 0.6497 | 0.5038 | 0.5106 | 0.4016 | 0.148 |
| timm_mobilevitv2_075 | image | 0.7319 | 0.5445 | 0.5424 | 0.2557 | 0.1384 |
| timm_densenet201 | image | 0.7133 | 0.5673 | 0.318 | 0.358 | 0.256 |
| torchvision_resnet101_imagenet1k_v2 | image | 0.7214 | 0.56 | 0.4893 | 0.5135 | -0.074 |
| timm_mobilevitv2_125 | image | 0.7384 | 0.5104 | 0.4966 | 0.3304 | 0.1315 |
| timm_dm_nfnet_f4 | image | 0.7683 | 0.4679 | 0.3572 | 0.4799 | 0.1328 |
| timm_mixer_l16_224_in21k | image | 0.7466 | 0.5074 | 0.3161 | 0.3135 | 0.3198 |
| timm_mixer_l16_224 | image | 0.7466 | 0.5074 | 0.3161 | 0.3135 | 0.3198 |
| torchvision_regnet_y_400mf_imagenet1k_v1 | image | 0.7413 | 0.4935 | 0.5198 | 0.3791 | 0.0667 |
| timm_dla102 | image | 0.7092 | 0.6133 | 0.3882 | 0.3459 | 0.1408 |
| torchvision_regnet_y_8gf_imagenet1k_v2 | image | 0.7012 | 0.4944 | 0.4507 | 0.4229 | 0.1281 |
| timm_convnext_femto_ols | image | 0.778 | 0.4716 | 0.3481 | 0.4967 | 0.0992 |
| timm_convnext_atto_ols | image | 0.7274 | 0.4354 | 0.5295 | 0.3632 | 0.1339 |
| timm_dla46_c | image | 0.8216 | 0.5016 | 0.4258 | 0.422 | 0.0136 |
| timm_hardcorenas_b | image | 0.7388 | 0.4791 | 0.5284 | 0.308 | 0.1228 |
| timm_gluon_resnet101_v1d | image | 0.6434 | 0.5673 | 0.2849 | 0.4331 | 0.2416 |
| torchvision_regnet_x_16gf_imagenet1k_v2 | image | 0.7278 | 0.5014 | 0.4388 | 0.3486 | 0.1519 |
| timm_gluon_resnet18_v1b | image | 0.752 | 0.5442 | 0.3602 | 0.3535 | 0.1578 |
| torchvision_regnet_x_3_2gf_imagenet1k_v1 | image | 0.7178 | 0.4522 | 0.3204 | 0.4663 | 0.2081 |
| timm_lcnet_100 | image | 0.7827 | 0.536 | 0.4504 | 0.2069 | 0.1881 |
| timm_mvitv2_base | image | 0.7543 | 0.5704 | 0.5032 | 0.4639 | -0.129 |
| timm_inception_v3 | image | 0.7661 | 0.4786 | 0.2779 | 0.4788 | 0.1603 |
| timm_mobilenetv3_rw | image | 0.7691 | 0.5035 | 0.3319 | 0.4195 | 0.1361 |
| torchvision_resnext101_32x8d_imagenet1k_v1 | image | 0.7762 | 0.5754 | 0.1969 | 0.3776 | 0.231 |
| timm_ecaresnet26t | image | 0.7326 | 0.5678 | 0.3292 | 0.4777 | 0.0486 |
| timm_cs3darknet_focus_m | image | 0.7961 | 0.5371 | 0.4674 | 0.4917 | -0.1446 |
| timm_legacy_seresnet34 | image | 0.7835 | 0.6036 | 0.3491 | 0.4289 | -0.019 |
| timm_densenet169 | image | 0.5511 | 0.5115 | 0.4031 | 0.4615 | 0.2091 |
| timm_efficientnet_b1 | image | 0.6615 | 0.4573 | 0.4815 | 0.4198 | 0.1113 |
| torchvision_mobilenet_v2_imagenet1k_v1 | image | 0.7786 | 0.4402 | 0.3555 | 0.426 | 0.1211 |
| timm_lambda_resnet26rpt_256 | image | 0.7237 | 0.6085 | 0.2321 | 0.4881 | 0.0657 |
| dino_vitb16 | image | 0.8148 | 0.548 | 0.0886 | 0.4409 | 0.2221 |
| timm_ecaresnet269d | image | 0.742 | 0.5016 | 0.5273 | 0.2732 | 0.069 |
| torchvision_densenet169_imagenet1k_v1 | image | 0.5514 | 0.5263 | 0.3956 | 0.4376 | 0.2022 |
| timm_dla34 | image | 0.7551 | 0.5184 | 0.4438 | 0.312 | 0.0829 |
| timm_lcnet_075 | image | 0.7438 | 0.5293 | 0.3299 | 0.2675 | 0.2405 |
| vissl_resnet50_clusterfit | image | 0.7686 | 0.5662 | 0.2448 | 0.3935 | 0.1299 |
| torchvision_densenet121_imagenet1k_v1 | image | 0.6889 | 0.5022 | 0.4292 | 0.358 | 0.1204 |
| timm_mobilenetv2_140 | image | 0.7526 | 0.4949 | 0.2789 | 0.4348 | 0.1327 |
| torchvision_regnet_y_8gf_imagenet1k_v1 | image | 0.7942 | 0.5928 | 0.3096 | 0.4036 | -0.0094 |
| timm_edgenext_xx_small | image | 0.7944 | 0.5335 | 0.5147 | 0.1278 | 0.1205 |
| timm_deit_tiny_patch16_224 | image | 0.7285 | 0.4965 | 0.4339 | 0.2595 | 0.1673 |
| timm_darknet53 | image | 0.6459 | 0.6054 | 0.2323 | 0.4527 | 0.1483 |
| timm_ecaresnet101d | image | 0.836 | 0.533 | 0.3541 | 0.223 | 0.1377 |
| timm_legacy_seresnext26_32x4d | image | 0.7734 | 0.562 | 0.3535 | 0.3331 | 0.0612 |
| timm_gluon_seresnext101_64x4d | image | 0.8212 | 0.604 | 0.3698 | 0.5227 | -0.2389 |
| timm_gluon_seresnext101_32x4d | image | 0.7768 | 0.4124 | 0.143 | 0.4789 | 0.263 |
| timm_lcnet_050 | image | 0.6679 | 0.4597 | 0.3826 | 0.2207 | 0.3386 |
| timm_dla60x_c | image | 0.7786 | 0.5474 | 0.2899 | 0.2251 | 0.2277 |
| timm_darknetaa53 | image | 0.7266 | 0.5523 | 0.2712 | 0.5034 | 0.0058 |
| torchvision_mnasnet0_5_imagenet1k_v1 | image | 0.7479 | 0.4795 | 0.4128 | 0.4231 | -0.0075 |
| timm_mnasnet_small | image | 0.6649 | 0.5395 | 0.607 | 0.0358 | 0.2076 |
| vissl_resnet50_jigsaw_goyal19 | image | 0.7013 | 0.5397 | 0.3567 | 0.4384 | 0.0126 |
| timm_mixnet_m | image | 0.7943 | 0.4956 | 0.3089 | 0.2048 | 0.2423 |
| vissl_resnet50_rotnet | image | 0.701 | 0.5713 | 0.1629 | 0.3847 | 0.2184 |
| timm_cs3sedarknet_l | image | 0.7197 | 0.4533 | 0.4444 | 0.3519 | 0.0681 |
| timm_convnext_femto | image | 0.7797 | 0.4834 | 0.1975 | 0.3765 | 0.181 |
| torchvision_resnet50_imagenet1k_v1 | image | 0.7322 | 0.477 | 0.28 | 0.3623 | 0.1555 |
| dino_xcit_small_12_p8 | image | 0.8229 | 0.6776 | 0.4106 | 0.3279 | -0.2367 |
| timm_jx_nest_small | image | 0.7662 | 0.6071 | 0.565 | 0.209 | -0.1565 |
| torchvision_regnet_y_32gf_imagenet1k_v1 | image | 0.7467 | 0.454 | 0.2998 | 0.1169 | 0.3332 |
| torchvision_densenet201_imagenet1k_v1 | image | 0.7144 | 0.5664 | 0.3134 | 0.3475 | 0.006 |
| torchvision_inception_v3_imagenet1k_v1 | image | 0.7665 | 0.4812 | 0.2544 | 0.2836 | 0.161 |
| timm_eca_nfnet_l2 | image | 0.7934 | 0.3318 | 0.186 | 0.2123 | 0.4089 |
| timm_lambda_resnet50ts | image | 0.7441 | 0.6037 | 0.1522 | 0.3766 | 0.042 |
| timm_gluon_resnet152_v1s | image | 0.6594 | 0.4908 | 0.3589 | 0.3501 | 0.041 |
| timm_mobilenetv3_small_100 | image | 0.732 | 0.574 | 0.4958 | -0.0076 | 0.0777 |
| timm_mobilevitv2_100 | image | 0.7801 | 0.5201 | 0.1054 | 0.3113 | 0.1447 |
| timm_mixnet_s | image | 0.7424 | 0.5093 | 0.3785 | 0.2781 | -0.0468 |

| timm_legacy_seresnext101_32x4d | image | 0.6634 | 0.4777 | 0.3745 | 0.1255 | 0.2168 |
|---|---|---|---|---|---|---|
| torchvision_squeezenet1_0_imagenet1k_v1 | image | 0.6452 | 0.4969 | 0.3821 | 0.171 | 0.1235 |
| timm_legacy_seresnet152 | image | 0.812 | 0.6272 | 0.0837 | 0.0999 | 0.1531 |
| timm_densenet161 | image | 0.666 | 0.4925 | 0.2929 | 0.3532 | -0.0611 |
| torchvision_squeezenet1_1_imagenet1k_v1 | image | 0.7709 | 0.429 | 0.2259 | 0.2374 | 0.0797 |
| torchvision_resnet152_imagenet1k_v2 | image | 0.6981 | 0.572 | 0.2959 | 0.2435 | -0.0836 |
| timm_cs3se_edgenet_x | image | 0.772 | 0.6367 | 0.0091 | 0.2194 | 0.0789 |
| torchvision_regnet_x_8gf_imagenet1k_v2 | image | 0.6967 | 0.5006 | 0.3604 | -0.0054 | 0.161 |
| c2d_r50 | video | 0.6384 | 0.5905 | 0.1191 | 0.1476 | 0.1294 |
| torchvision_regnet_y_1_6gf_imagenet1k_v1 | image | 0.7651 | 0.4819 | 0.429 | 0.2633 | -0.3227 |
| timm_efficientformer_l3 | image | 0.8125 | 0.5459 | 0.0514 | -0.132 | 0.1477 |
| torchvision_densenet161_imagenet1k_v1 | image | 0.6656 | 0.493 | 0.3019 | -0.0464 | -0.0809 |

**Supplemental table 1: List of vision DNNs tested.** This table provides a complete list of all pretrained image and video deep neural networks included in our analysis, ranked by their performance on the average score across five ratings. The naming convention is from DeepJuice (Conwell et al., 2024).